\documentclass[final,5p,times,twocolumn]{elsarticle}

\usepackage{amssymb}
\usepackage{url}
\usepackage{graphicx}
\usepackage{CJKutf8}
\usepackage{multirow}
\usepackage{booktabs}
\usepackage{hhline}
\usepackage{stfloats}
\usepackage{txfonts}
\usepackage{amsthm,amsmath,amssymb}
\usepackage{mathrsfs}
\usepackage{float}
\usepackage{balance}
\usepackage{subcaption}
\usepackage{color}
\usepackage{graphicx, amsmath, amssymb, caption, subcaption, multirow, overpic, textpos}
\usepackage{balance}
\usepackage{hyperref}
\usepackage{cleveref}

\hypersetup{hypertex=true,
            colorlinks=true,
            linkcolor=blue,
            anchorcolor=blue,
            citecolor=blue}
\usepackage{caption, subcaption}
 
\journal{Neurocomputing}

\begin{document}

\begin{frontmatter}



\title{Stroke-Based Autoencoders: Self-Supervised Learners for Efficient Zero-Shot Chinese Character Recognition}


\author[a]{Zongze Chen}
\author[a]{Wenxia Yang\corref{cor1}}
\cortext[cor1]{Corresponding author at: Dept. of Mathematics, Wuhan University of Technology, 
Wuhan 430070, China}
\ead{wenxiayang@whut.edu.cn}
\author[b]{Xin Li}

\affiliation[a]{organization={Dept. of Mathematics, Wuhan University of Technology, 
Wuhan 430070, China}}
\affiliation[b]{organization={Lane Dept. of Computer Science and Electrical Engineering, West Virginia University, Morgantown 26506, USA}}

\begin{abstract}
Chinese characters carry a wealth of morphological and semantic information; therefore, the semantic enhancement of the morphology of Chinese characters has drawn significant attention. 
The previous methods were intended to directly extract information from a whole Chinese character image, which usually cannot capture both global and local information simultaneously. 
In this paper, we develop a stroke-based autoencoder(SAE), to model the sophisticated morphology of Chinese characters with the self-supervised method.
Following its canonical writing order, we first represent a Chinese character as a series of stroke images with a fixed writing order, and then our SAE model is trained to reconstruct this stroke image sequence. This pre-trained SAE model can predict the stroke image series for unseen characters, as long as their strokes or radicals appeared in the training set. 
We have designed two contrasting SAE architectures on different forms of stroke images. One is fine-tuned on existing stroke-based method for zero-shot recognition of handwritten Chinese characters, and the other is applied to enrich the Chinese word embeddings from their morphological features.
The experimental results validate that after pre-training, our SAE architecture outperforms other existing methods in zero-shot recognition and enhances the representation of Chinese characters with their abundant morphological and semantic information.
\end{abstract}






\end{frontmatter}


\section{Introduction}

Natural language processing (NLP) tasks have achieved great success using self-supervised learning methods. Large-scale pre-trained models, such as Bert\cite{devlin2018bert} and GPT\cite{radford2019language}, have become essential backbones in text classification\cite{minaee2021deep}, question answering\cite{bordes2015large}, and natural language understanding\cite{liu2019roberta}. 
Chinese characters emerged as hieroglyphs and naturally carry very rich morphological and semantic information. 
Therefore, the pictographic structures of Chinese characters can further facilitate many language-dependent NLP tasks, such as the recognition of Chinese names\cite{wu2021mect, Li2021MFENERMF} and the segmentation of Chinese words\cite{Sun2021ChineseBERTCP}.
For semantic enhancement, artificial encoding and image-based representation are the two main encoding styles for Chinese characters. Character encoding methods are often at Radical-level\cite{Li2021MFENERMF} or Stroke-level\cite{cw2vec}.  But these human-generated encoding approaches carry no spatial information and may lead to repetition. 
For image representation, most methods extract information from Chinese character pictures directly\cite{Sun2021ChineseBERTCP, Li2021GlyphCRMBE}, which generally fails to utilize global information with local information of Chinese characters simultaneously. 
Based on the above analysis, we explore a self-supervised learning strategy on Chinese character images to learn the encoding of Chinese characters and attempt to extract the characters' semantic information in a more comprehensive and effective style.
\begin{figure}[ht]
  \centering
  \center{\includegraphics[width=8cm]  {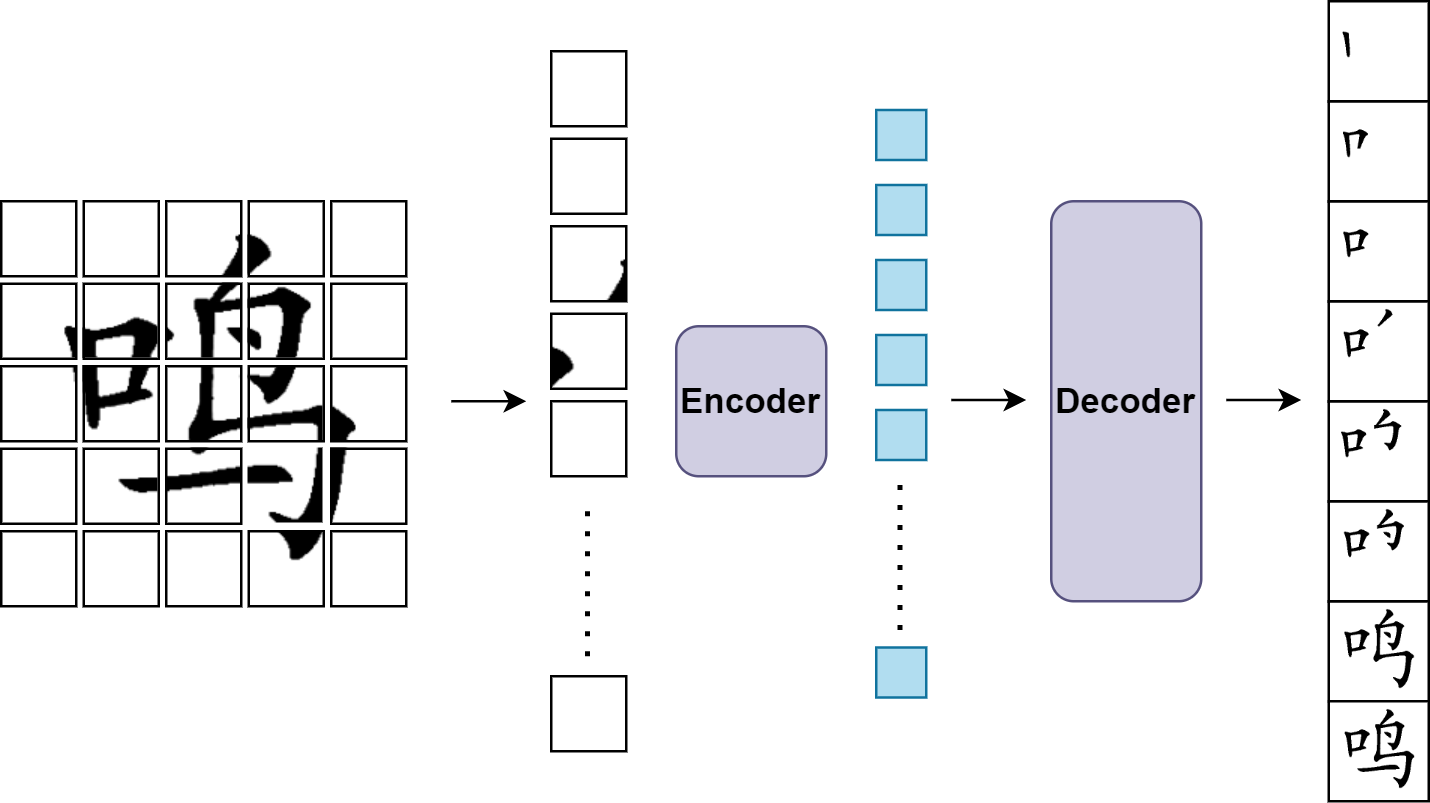}}
  \caption{\textbf{Our SAE architecture based on Vision Transformer.} The encoder is applied on the image patches of the Chinese characters (we take  '\begin{CJK*}{UTF8}{gbsn}鸣\end{CJK*}' for example).  The decoder reconstructs the stroke sequence images(the order of strokes) from the encoded patches.}
  \label{fig:1}
\end{figure}

Self-supervised learning on images learns image representations by recovering masked images based on the encoding of input images.
Inspired by Transformer, Dosovitskiy et al. \cite{Dosovitskiy2021AnII} apply Transformer directly to the sequences of image patches for image recognition, and Bao et al. \cite{Bao2021BEiTBP} propose a masked image modeling task to pre-train Vision Transformers. 
Most recently, masked aotuencoder (MAE) \cite{He2021MaskedAA} masks a high proportion(75\%) of the input images and performs well on downstream tasks, such as image recognition, object detection, instance segmentation, to name a few.
However, these methods cannot be applied directly to Chinese character images, since there are great differences between Chinese character images and general images in terms of both image structure features and semantic information.
To be specific, \textbf{(i)} Chinese character images are usually space discrete, with the smallest unit stroke, while the general images are continuous with unit pixel, so the random mask may not work. \textbf{(ii)} Chinese characters are human-generated signals that are information dense with highly structured semantics, and every stroke or radical part may have its special informative meaning. \textbf{(iii)} Each Chinese character can be represented as a time series of images that carry information throughout its fixed order of strokes, as shown in Fig. \ref{fig:1}. 

\begin{figure*}[ht]
  \centering
  \center{\includegraphics[width=15cm]  {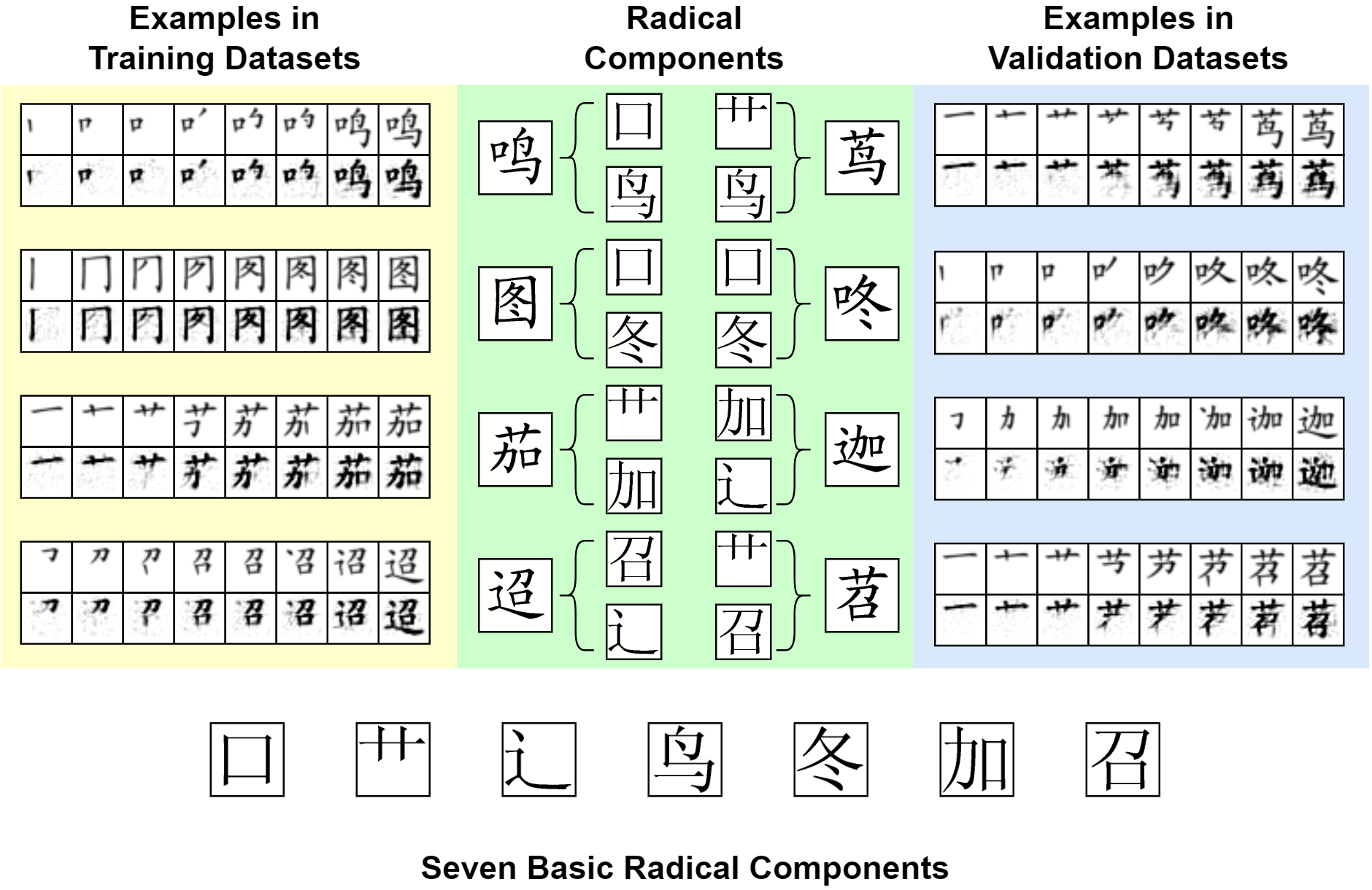}}
  \caption{Example results on training(seen class) and validation(unseen class) datasets.  Each character has the original picture(above) and reconstructed picture(below). The left column is the Chinese character image in the training datasets, and the right column is the Chinese character image in the validation datasets. The four characters of each column consist of seven basic radical parts.}
  \label{fig:2}
\end{figure*}
Driven by the above analysis, we propose a stroke-based autoencoder (SAE) to extract the morphological information from the sequences of each Chinese character images. More specifically, our framework consists of two parts: \textbf{(i)} the encoder network, which operates on the image patches of Chinese characters to obtain encoded patches, and \textbf{(ii)} the decoder network, which reconstructs the stroke sequence images of each corresponding Chinese character. The encoder in our SAE pre-training model aims at extracting the semantic information from stroke images, and the decoder learns to predict the sequences of the writing order of Chinese characters.
Some example results are illustrated in Fig. \ref{fig:2}. We choose some Chinese characters, all of which have eight strokes for a neat illustration.  The left column (with a yellow background) is the images of Chinese characters \begin{CJK*}{UTF8}{gbsn}'鸣', '图', '茄' and '迢'\end{CJK*} in training dataset, and the right column (with a pale blue background)  are images of \begin{CJK*}{UTF8}{gbsn}'茑', '咚', '迦' and '苕'\end{CJK*} in the validation data set. Each Chinese character can be divided into two radical components, as illustrated in the middle column (with a light green background).  These characters are made up of seven basic radical components, as illustrated at the bottom of the figure. 
For each character, the first row is the original stroke sequence images, i.e., their canonical order of strokes, and the second row is the corresponding stroke sequence predicted by our SAE model.
The example results demonstrate that our pre-training model can successfully predict and reconstruct the unseen strokes and radicals of Chinese characters.

Therefore, Our SAE can be applied to enrich word embeddings for downstream NLP tasks such as Chinese Named Entity Recognition. It can also be applied to explore the application of Chinese character structures, such as zero-shot Chinese character recognition(zero-shot CCR).
In this paper, we focus on the morphological structures of Chinese characters and utilize our pre-training models to explore zero-shot Chinese character recognition and word embeddings. we design two different SAE architectures on different forms of stroke images for different tasks. 
The experimental results validate that our SAE outperforms the previous results\cite{Chen2021ZeroShotCC, cao2020zero, wang2018denseran} on zero-shot recognition of Chinese characters. What is more encouraging is that our SAE is pre-trained on a small and concise dataset with printed Chinese character images, but boosts huge performance in zero-shot recognition on a large and complicated dataset consisting of handwritten Chinese character images.
Meanwhile, the word embedding generated by our SAE is used to calculate the word similarity of Chinese characters and their radical parts as an intrinsic task. The experiment shows that our SAE can reliably extract rich semantic information from Chinese characters.

The main contributions of our work are as follows.

\textbf{(1)} We present a stroke-based autoencoder (SAE) pre-training model through self-supervised learning. The pre-trained SAE model successfully captures the morphological and semantic information of Chinese characters.

\textbf{(2)} We design two contrasting architecturesc(Vision Transformer vs. Resnet+Transformer)  in varying forms of stroke images for different tasks, namely zero-shot Chinese character recognition and semantic enhancement.

\textbf{(3)} The pre-trained SAE model is fine-tuned with the existing stroke-based method and outperforms other existing methods in zero-shot recognition. This model can also be applied to enrich Chinese word embeddings from their morphological structures.

\section{Related work}
\begin{figure}[h]
  \centering
  \center{\includegraphics[width=8.3cm]  {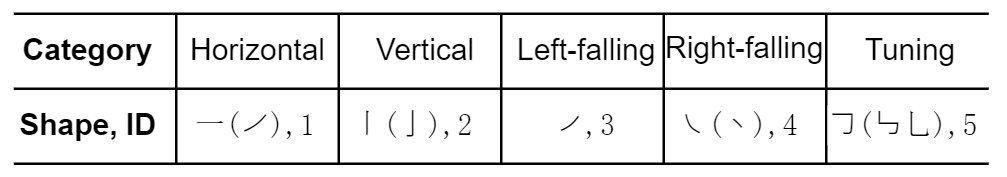}}
  \caption{ Five basic categories of strokes.}
  \label{fig:3}
\end{figure}

\noindent
\textbf{Natural Language Processing} Deep learning-based neural networks are the prevailing models applied in almost all NLP tasks. Recurrent sequence modelings, such as RNN\cite{zaremba2014recurrent}, LSTM\cite{hochreiter1997long}, and Transformer\cite{Vaswani2017AttentionIA}, predict the next word based on previous information. Masked language modeling, such as BERT\cite{devlin2018bert} and GPT\cite{radford2019language}, has been a great success.  By training the model with a portion of the input sequence, these models learn the representation of the word and then predict the missing parts. The latest research suggests that pre-training methods \cite{devlin2018bert,Radford2018ImprovingLU, Brown2020LanguageMA} generalize well to various downstream tasks.
\\
\textbf{Masked Image Encoding} To learn image representations by predicting masked images from an input sequence of images, DAE\cite{Vincent2010StackedDA} first considers masks as noises and obtain image representation by denosing the corrupted input signals. This pioneering work is generalized to a number of masked image encoding methods. Inspired by the Transformer, many approaches \cite{Bao2021BEiTBP,Chen2020GenerativePF,Dosovitskiy2021AnII} are proposed for image encoding. The Vision Transformer(ViT)\cite{Dosovitskiy2021AnII} performs a preliminary exploration of masked patch prediction for self-supervised learning. Beit\cite{Bao2021BEiTBP} predicts the discrete tokens\cite{Ramesh2021ZeroShotTG} to pre-train ViT. Most recently, MAE\cite{He2021MaskedAA} masks about 75\% proportion of the input image and achieves better recognition accuracy on ImageNet-1K and other datasets.
\\
\textbf{Semantic Enhancement from the morphology of Chinese Characters} Cao et al. \cite{cw2vec} first exploit the stroke-level information of Chinese characters with stroke n-grams. Li et al.\cite{Li2021MFENERMF} encode Chinese characters with Five-Stroke for the recognition of Chinese Named Entities. But these human-generated encoding methods may lead to repetition, since they cannot take full advantage of the spatial information of Chinese characters.
Instead of using the ID-based character embedding method, Sun et al.\cite{Sun2021ChineseBERTCP} and Li et al.\cite{Li2021GlyphCRMBE}  extract the glyph information of Chinese characters for the language model based on Chinese character images. However, due to the lack of supervision, these methods cannot focus on global and local information simultaneously. 
In addition, word embedding evaluation plays a important role in the estimation of semantic enhancement. Evaluation methods can be roughly divided into two categories: Intrinsic evaluation with word similarity\cite{camacho2017semeval} or word analogy\cite{li2018analogical}, and extrinsic evaluation for the performance of downstream NLP tasks. It has been verified that there are strong correlations between intrinsic and extrinsic evaluations\cite{qiu2018revisiting}.\\
\textbf{Zero-Shot Chinese Character Recognition} Zero-Shot Chinese character recognition depends on the set of labeled training set of seen classes and some preliminary knowledge about the semantic relation between unseen classes and seen classes\cite{Kodirov_2017_CVPR}.
This semantic relation is often described as a high-dimensional vector , which is established from a semantic embedding space (e.g., the attribute space\cite{6571196}).
Specifically, most existing zero-shot learning methods learn a projection function from the visual feature space to the semantic embedding space\cite{Kodirov_2017_CVPR}, by training visual data with labels consisting only of visible classes.
For zero-shot Chinese character recognition, the semantic embedding space is heavily based on the exploration of the structures of Chinese characters, which can be explored from radical-level\cite{wang2018denseran, wang2019radical, zhang2018radical} and stroke-level\cite{Chen2021ZeroShotCC}.
Cw2vec\cite{cw2vec} first exploits the stroke-level information of Chinese characters with stroke n-grams, which divides Chinese characters into five basic categories of strokes such as horizontal, vertical, left-falling, right-falling, and turning, according to the Chinese national standard GB18030-20, as shown in Fig. \ref{fig:3}. For example, the Chinese character '\begin{CJK*}{UTF8}{gbsn}鸣\end{CJK*}' can be encoded as '25135451' as shown in Fig. \ref{fig:4}. Chen et al.\cite{Chen2021ZeroShotCC} then used this stroke-based encoding as the semantic embedding of the Chinese character, making breakthrough progress in zero-shot recognition.
However, this human-generated encoding still lacks spatial information and has the potential for repetition.

\section{Prior structural Knowledge about Chinese Characters}
In this section, we provide some prior knowledge of the structures of Chinese characters to help us better understand Chinese characters.

\begin{figure}[ht]
  \centering
  \center{\includegraphics[width=8cm]  {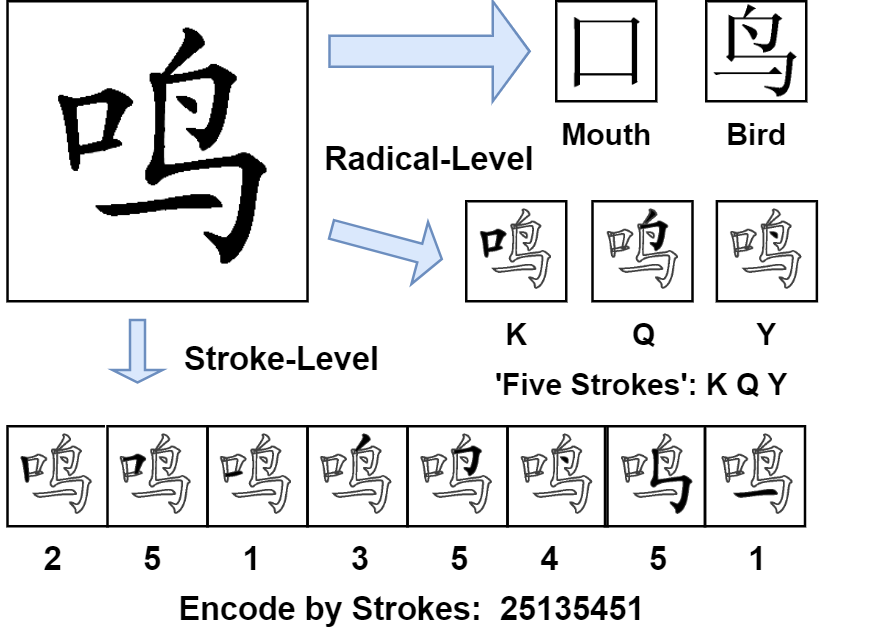}}
  \caption{Encoding methods in radical-level and stroke-level}
  \label{fig:4}
\end{figure}

Chinese characters are hieroglyphs, which can be divided into radical level or stroke level\cite{Chen2021ZeroShotCC}. At the radical level, Chinese characters with similar structures often have similar meanings. As shown in Fig.\ref{fig:4}, the meaning of the Chinese character '\begin{CJK*}{UTF8}{gbsn}鸣\end{CJK*}'  is 'the chirping of birds', and '\begin{CJK*}{UTF8}{gbsn}鸣\end{CJK*}'  can be split into two radicals: '\begin{CJK*}{UTF8}{gbsn}口\end{CJK*}'(month)  and'\begin{CJK*}{UTF8}{gbsn}鸟\end{CJK*}'(bird). Thus, Chinese characters are naturally informative datasets\cite{enriching, Liu2010HolisticVA}. This abundant information is often used to enrich Chinese word embeddings in NLP tasks\cite{cw2vec}. Meanwhile, according to the Five Strokes input method, most Chinese characters can be divided into no more than four basic Chinese character structures. Based on the 'Five Strokes input method', GlyphCRM\cite{Li2021MFENERMF} provides a bidirectional encoding method for Chinese characters with the Glyph. As shown in Fig. \ref{fig:4}, '\begin{CJK*}{UTF8}{gbsn}
鸣\end{CJK*}' also can be encoded as 'KQY' using the 'Five Strokes input method'.

At stroke level, Chinese characters can be divided into five basic categories of strokes\cite{cw2vec}: horizontal, vertical, left falling, right falling, and turning according to the Chinese national standard GB18030-2005, as shown in Fig. \ref{fig:3}.
Compared to radical level, the stroke-level based presentation has superiority of conciseness and consistency, while radical level is more complex with less space effectiveness.
Moreover, Chinese characters have some general writing principles such as "from left to right", "top to bottom", "inside to outside". With this principle,  '\begin{CJK*}{UTF8}{gbsn}鸣\end{CJK*}' can be encoded by strokes as ‘25135451', as illustrated in Fig. \ref{fig:4}. This property reminds us that when the model learns the orders of strokes of some characters, it can naturally master how to write new characters, even if these characters have never appeared in the training set. 
\section{Approach}
In this section, we introduce the proposed SAE architectures and the pre-training model. We develop two contrasting architectures for different tasks. For zero-shot Chinese character recognition, we propose an SAE-based on Resnet and Transformer, while for semantic enhancement, we propose an SAE-based on Vision Transformer.  In the experimental part, we will explain why these two models match with different Chinese NLP tasks.

\subsection{The Pre-training model}
\begin{figure}[h]
  \centering
  \center{\includegraphics[width=8cm]  {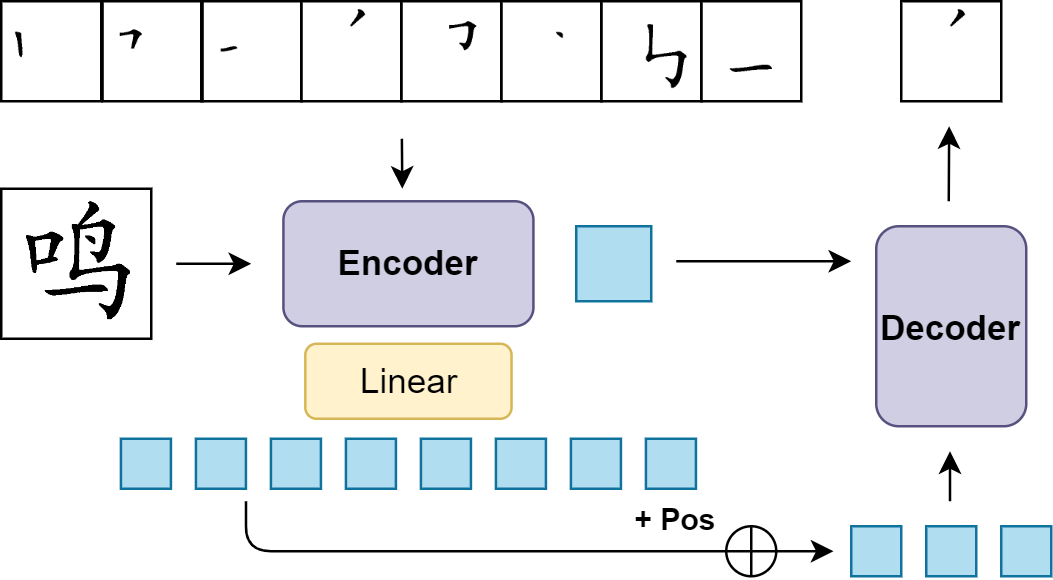}}
  \caption{\textbf{Our SAE architecture based on Resnet and Transformer.} The encoder is applied on the images of Chinese characters and their image sequence of labels. The deocder reconstructs the stroke sequence images through the encoded patches of original image and stroke sequence images at the previous moment.}
  \label{fig:5}
\end{figure}

\begin{figure}[h]
  \centering
  \center{\includegraphics[width=8cm]  {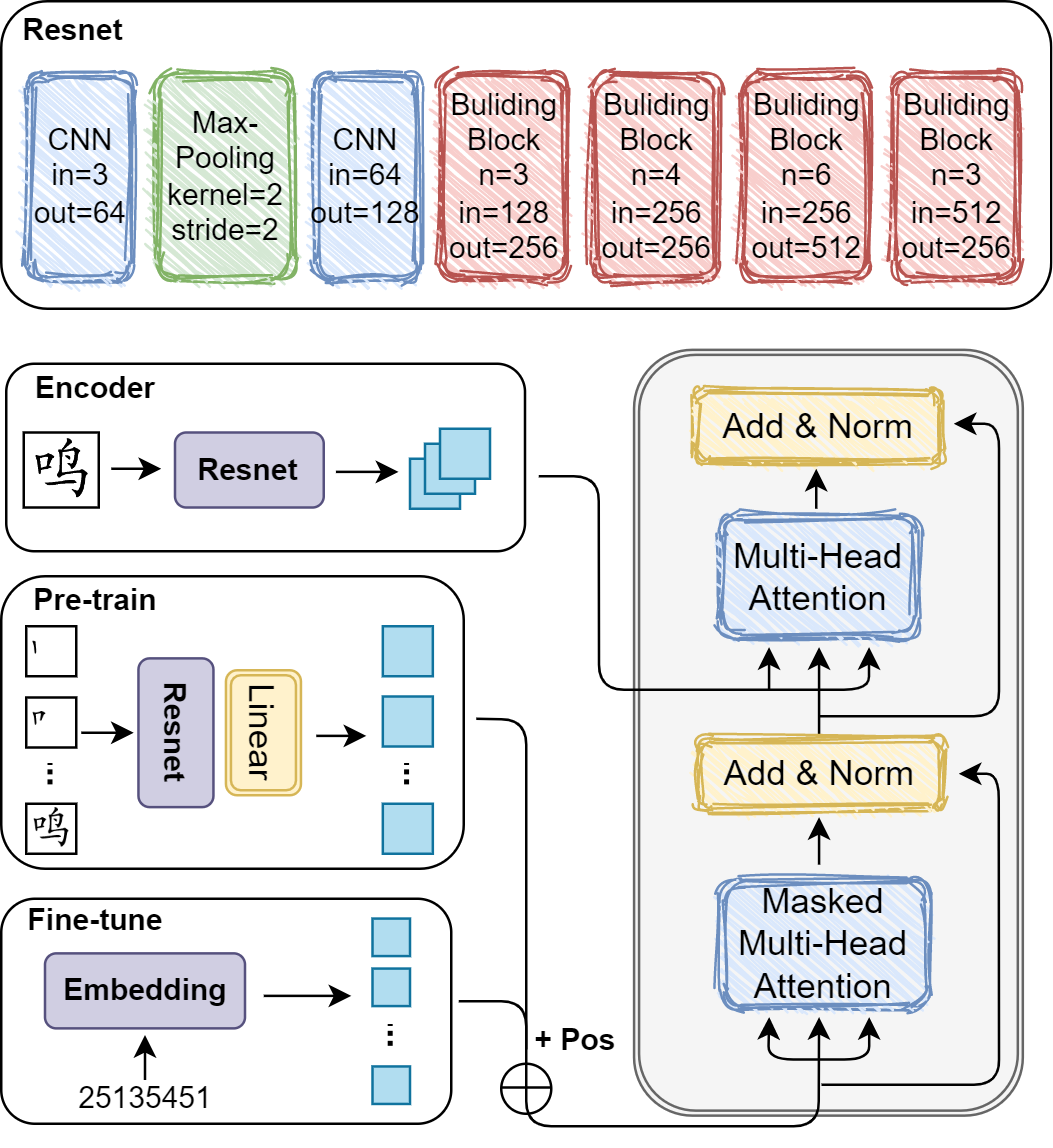}}
  \caption{The detailed architecture of encoder and decoder.}
  \label{fig:6}
\end{figure}

Our stroke-based autoencoder(SAE) is a self-supervised model that predicts the original signal given its encoded patches, as shown in Fig. \ref{fig:1} and Fig. \ref{fig:5}. Similar to other autoencoders, our model has an encoder that maps an input sequence of symbol representations $x = (x_1, ..., x_n), x\in \mathbb{R}^{n \times  d_1}$ to a sequence of continuous representations $y = (y_1, ..., y_n)\in \mathbb{R}^{n\times d_2}$,  and a decoder that generates an output sequence $z = (z_1, ..., z_n)\in \mathbb{R}^{n\times d_3}$. where $d_1$ 
is the number of pixels in the input patches, $d_3$ is the dimension of the outputs and $d_2$ is the dimension embedded of the encoded patches.
Unlike the Masked Autoencoder\cite{He2021MaskedAA}, which uses the spatial segmentation of pictures, we divide Chinese pictures by time according to their writing orders, since every Chinese character has its unique writing order. In this way, we reconstruct the stroke sequence of character images from the encoded patches.\\
\textbf{SAE based on Vision Transformer}. Our encoder based on  Vision Transformer(ViT) \cite{Dosovitskiy2021AnII} is applied to the image patches of a Chinese character picture, as shown in Fig. \ref{fig:4}. For a Chinese character, we set its stroke number to $m$. In our Chinese character datasets, the maximum stroke number of characters is 24. Thus, we split each image into $n = 5 \times 5$ patches $x = (x_1, ..., x_n)$. For these characters whose stroke number is less than 25, we supplement their stroke sequence images with blank pictures.
We apply a linear projection(Proj) with added positional embeddings(Pos) on the input patches; after that, a series of Transformer blocks are used to process the encoded patches. The encoder has depth 4 and width 256 of Transformer blocks. 
\begin{align}
    y & = Transformer\_Blocks\_1(Proj(x) + Pos)\\
    y & = Transformer\_Blocks\_i(y), i= 2,3,4
\end{align}
 Our decoder is similar to the decoder of Transformer\cite{Vaswani2017AttentionIA} applied on encoded patches, but only consists of multihead self-attention. That is, we directly reconstruct the image sequence of a Chinese character with encoded patches. The decoder also has depth 4 and width 256 of the Transformer blocks.
\begin{align}
    z & = Transformer\_Blocks\_1(y + Pos)\\
    z & = Transformer\_Blocks\_j(z)), j = 2,3,4
\end{align}
\\
\textbf{SAE based on Resnet and Transformer} If we consider the stroke sequence of images as a sentence, each image can be viewed as a word. Thus, the reconstruction task can be modeled as a sequence-to-sequence (seq2seq) task. In this regard, we exploit the Transformer\cite{Vaswani2017AttentionIA}to decode the stroke sequence of images from the previous images and the original image of a Chinese character.
Considering that Resnet\cite{He2016DeepRL} plays an important role in recognition tasks and performs better on smaller datasets than Vision Transformer, we propose an SAE architecture based on Resnet and Transformer, the overall framework is illustrated in Fig. \ref{fig:5}. For an input image $x\in \mathbb{R}^{C\times H \times W}$,  we employ Resnet as the encoder to output $\frac{H}{2} \times \frac{W}{2} \times 256$ encoded patches. For input image sequences, a full connection layer is added after Resnet to map $\frac{H}{2} \times \frac{W}{2} \times 256$ to $1 \times 256$ word vectors.
The encoder here has the effect of generating word vectors from image sequences. The decoder has depth 1 and width 256 of Transformer blocks. 
The detailed architecture of our encoder and decoder is shown in Fig. \ref{fig:6}.
\\
\textbf{Reconstruction target}. In the pre-training stage, our SAE reconstructs the image by predicting the pixel values of the encoded patches. The last layer of the decoder is a full connection layer whose number of output channels is equal to the number of pixels in each patch. The output of the decoder is then reshaped into a reconstructed image. The reconstruction loss function computes the mean squared error(MSE) between the reconstructed and original images in pixel space.
\subsection{Fine-tuning  for zero-shot Chinese Characters Recognition }
\begin{figure}[h]
  \centering
  \center{\includegraphics[width=8cm]  {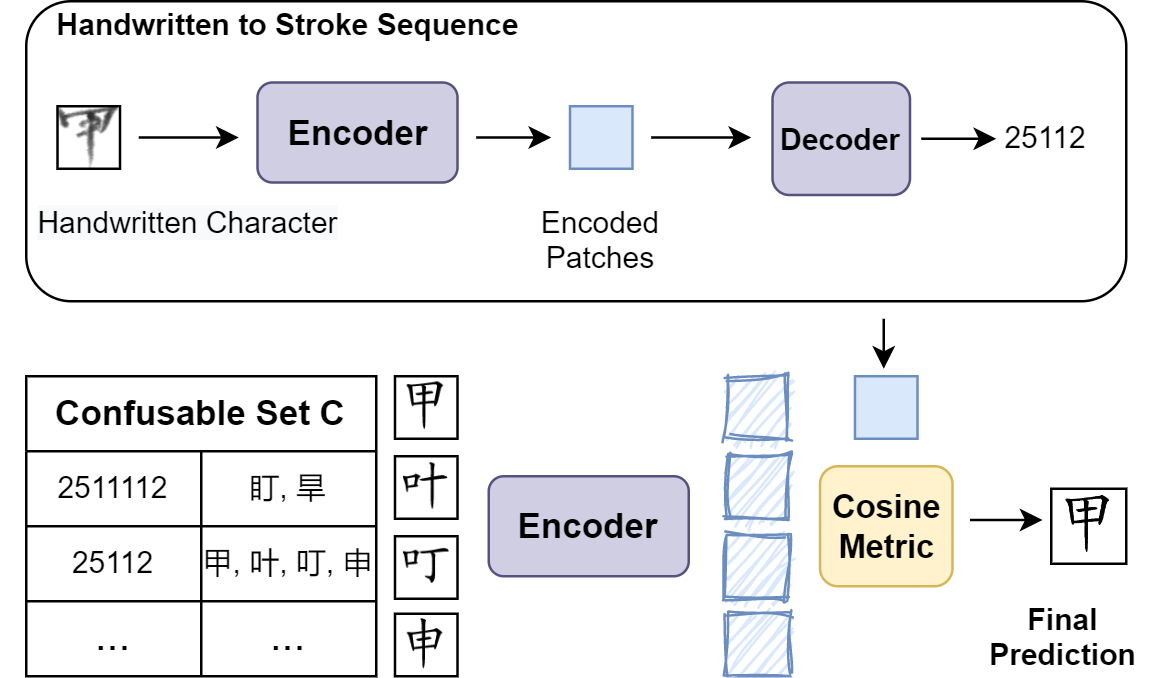}}
  \caption{The architecture of the stroke-based zero-shot recognition method.}
  \label{fig:7}
\end{figure}

Chen et al. \cite{Chen2021ZeroShotCC} propose a stroke-based method for zero-shot Chinese character recognition, which makes breakthrough progress in zero-shot recognition. This method uses ResNet\cite{He2016DeepRL} as the Image-to-Feature Encoder and uses a series of Transformer blocks\cite{Vaswani2017AttentionIA} as the Feature-to-Stroke Decoder.
We fine-tune our pre-trained model on this method for zero-shot Chinese character recognition. 
The basic architecture is similar to the stroke-based autoencoders illustrated in Fig. \ref{fig:5}. The ground-truth is denoted by $t = (t_1,t_2,...,t_n)$ , and the Cross-Entropy loss is used to optimize the model: $loss = -\sum_{i}^{n}{log(p(t_i)}$, where $p(t_i)$ is the probability of class $t_i$ at the $i-th$ time step.
Besides, we notice that different Chinese characters may have the same stroke sequences, that is, one label may correspond to multiple Chinese characters. As Fig. \ref{fig:7} illustrates, the four characters \begin{CJK*}{UTF8}{gbsn}'甲', '叶', '叮' and '申'\end{CJK*}  have the same stroke sequences. To tackle this one-to-many problem, Chen constructs the confusable set $C$, which includes these characters that have the same stroke sequence, and then feeds both the input handwritten character image and the printed images in set $C$ into the encoder. After getting the encoded patches of the input image(denoted by $F$) and the encoded patches of printed images in $C$ (denoted by $\mathcal{F^{'}}= \{F^{'}_1, F^{'}_2, ..., F^{'}_N\}$ ), the similarity between $F$ and  $F^{'}_i$ is calculated, and the final Chinese character is determined by Eq. \eqref{eq:5}. 
\begin{equation}
    i^* = 
    \mathop {\arg \max }\limits_{i\in\{1,2,...,N\}}
    \frac{F^TF^{'}_i}
    {\|F\|\times\|F^{'}_i\|}
    \label{eq:5}
\end{equation}

For our zero-shot recognition task, we apply the SAE based on ResNet and Transformer to predict the stroke sequences of Chinese characters after pre-training. The last layer of the decoder is a linear projection whose number of output channels changes from the number of pixel values to the number of categories. The loss function for fine-tuning  computes the cross entropy loss between the prediction and the ground truth.

For a general transfer learning task, it is essential to pre-train the model on a large dataset to obtain the abundant knowledge. Then the pre-trained model can generalize well to the target task with a limited dataset.
For our zero-shot tasks, however, pre-training datasets are much smaller than fine-tuning datasets, besides, the handwritten Chinese characters images from fine-tuning datasets are more complicated than printed Chinese character images in the pre-training datasets.
As a result, the model parameters learned by pre-training may not generalize well in our zero-shot tasks.

To fix this problem, we need to figure out the respective strengths of image reconstruction and zero-shot recognition, and explore how to combine them. Our SAE architecture has a better generalization performance in stroke sequence prediction by reconstructing the stroke sequence of images. This performance is learned from the deep layer of the encoder and decoder. The zero-shot recognition task teaches the model to extract rich and more generalized features from large and complicated datasets, i.e. handwritten Chinese character images. This performance is learned from the shallow layer of the encoder.

We present a new fine-tuning method by combining the training parameters learned from handwritten Chinese character images and pre-training parameters learned from printed Chinese character images. We first train the stroke-based model\cite{Chen2021ZeroShotCC} on zero-shot recognition tasks, and overwrite the parameters of the last building block of the encoder along with the parameters of the decoder with our pre-trained parameters.
Then, we freeze the parameters of the first three building blocks of the encoder, namely the parameters learned from handwritten Chinese character images. The rest of the parameters are fine-tuned for zero-shot tasks in the end.

\section{Experiments}
To evaluate the effectiveness of our proposed SAE model, we carry out comprehensive experiments on pre-training and Zero-Shot Chinese character recognition. We demonstrate both qualitative and quantitative evaluations of the performance of our approach. 

\begin{figure}[h]
  \centering
  \center{\includegraphics[width=8cm]  {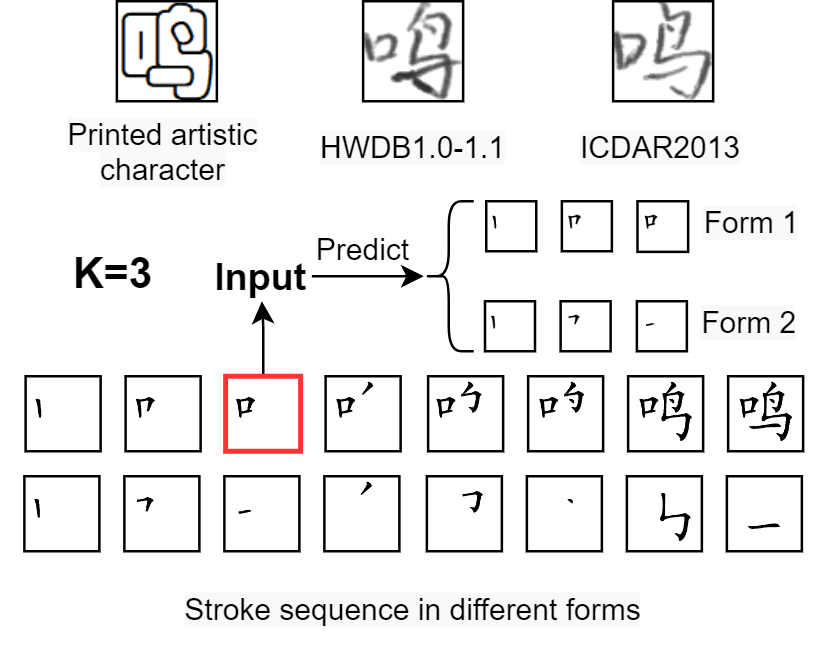}}
  \caption{Examples of datasets. The last two lines of images are different forms of Chinese stroke sequences.}
  \label{fig:8}
\end{figure}
\subsection{Datasets} 

For the pre-training procedure, we provide two forms of stroke sequence of printed Chinese characters. As shown in Fig. \ref{fig:8}, for the set of sequences of stroke images in the first form, each subsequent stroke image contains the previous stroke information of a Chinese character. For the second form, each subsequent stroke image contains only the individual strokes.
For each form, we collected 36,670 stroke sequence images of 3,755 printed Chinese characters from Hanzi Writer \cite{hanziwriter}. Each stroke corresponds to an image.  For the reconstruction task, the first 500, 1000, 1500, 2000 and 2755 classes of 3755 printed Chinese character images are used as the training sets, and the last 1000 classes the validation set.
Furthermore, as marked in Fig. \ref{fig:8}, the third stroke image is a Chinese character '\begin{CJK*}{UTF8}{gbsn}口\end{CJK*}', which contains the stroke information of the first three images. Thus, we can take the third stroke image as the input of SAE and predict the first three stroke images. We augment the datasets by taking the \textit{k}-th stroke images to predict previous stroke images, where $k \in \{1,2,...,m\}$. In this way, we can improve the generalization performance of our pre-training model. 

For the fine-tuning procedure, we use HWDB1.0-1.1 \cite{Liu2013OnlineAO} as the training set, which contains 2,678,424 offline handwritten Chinese character images with 3,881 classes collected from 720 writers.
We use ICDAR2013\cite{Yin2013ICDAR2C} as the validation set to evaluate the performance of our model. There are totally 224,419 offline handwritten Chinese character images and 3,755 classes collected from 60 writers in ICDAR2013.
We also used the 105 printed artistic fonts for 3,755 characters (394,275 samples) collected by Chen et al.\cite{Chen2021ZeroShotCC} for this recognition task. 
For the zero-shot recognition task, we choose samples from the first 500, 1000, 1500, 2000 and 2755 classes as the training set and the last 1000 classes as the test set. 

\subsection{Implementation Details}
Our model is implemented with PyTorch and the experiments are carried out on a server with 4 NVIDIA RTX2080 GPU with 8GB memory. AdamW\cite{adamw} optimizer is used with the initial learning rate set to 1e-4. And the cosine annealing rule\cite{loshchilov2016sgdr}: $lr_t = \eta_min + (lr_max - lr_min)*(1+cos\frac{T_{cur}}{T}\pi)$ is used to decay the learning rate, where $T_{cur}$ and $T$ represent the current epoch and the total number of epochs, respectively. The batch size is set to 32. The optimizer momentums $\beta_1$ and $\beta_2$ are set to 0.9 and 0.999, the weight decay is set to 0.05 empirically.

For ViT-based pre-training in SAE, the input images are resized to $140\times140$ and the label images are resized to $28\times28$. 
The input images of SAE based on Resnet and Transformer are resized to $28\times28$. Grayscale images are normalized to [0,1]. 
\begin{figure}[h]
  \centering
  \center{\includegraphics[width=8cm]  {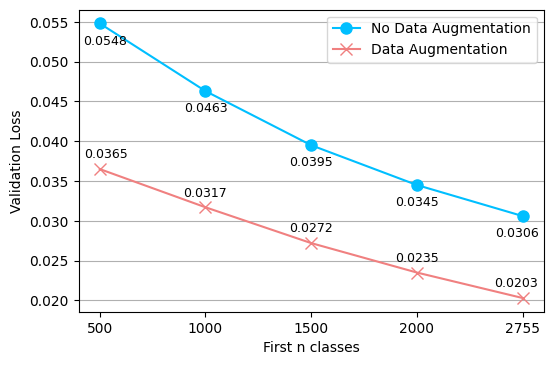}}
  \caption{Validation loss in different data sets and data augmentation. The x-axis is the number of sample classes in the training dataset, i.e. first 500,1000,1500,2000,2755 classes. The y-axis is the MSE loss on last 1000 classes.}
  \label{fig:9}
\end{figure}

Data augmentation is vital for pre-training on small datasets. As shown in Fig. \ref{fig:2}, for the characters '\begin{CJK*}{UTF8}{gbsn}图\end{CJK*}' and '\begin{CJK*}{UTF8}{gbsn}咚\end{CJK*}', they are both composed of the same size of '\begin{CJK*}{UTF8}{gbsn}冬\end{CJK*}' and different size of '\begin{CJK*}{UTF8}{gbsn}口\end{CJK*}'. Thus, it is important to generalize across glyphs of different sizes in a small dataset. We use random resized cropping by setting a random area scale between 0.8 and 1, and a random aspect ratio between $\frac{3}{4}$ and $\frac{4}{3}$ on the same set of samples and labels.
Fig. \ref{fig:9} shows the validation loss in different data sets with and without data augmentation. It can be seen that the validation loss has been dramatically reduced by data augmentation.
For fine-tuning in our zero-shot task, no other data augmentation strategies are used, since our dataset is large enough.

\begin{table*}[ht]
\caption{The results of zero-shot recognition with various approaches. Our model utilize the SAE based on ResNet and Transformer pre-trained on the second form of data sets.}
\begin{center}
\setlength{\tabcolsep}{1mm}{
\begin{tabular}{cccccc|cccccc}
\hline
{\multirow{2}{*}{\centering Handwritten}} & \multicolumn{5}{c|}{CCR Accuracy under Different Seen Classes} & \multirow{2}{*}{\centering Printed Artistic} & \multicolumn{5}{c}{CCR Accuracy under Different Seen Classes} \\ 
\hhline{~-----~-----}
&  500 & 1000 & 1500 & 2000 & 2755  &    &500 & 1000 & 1500 & 2000 & 2755 \\ \hline
                                      DenseRAN\cite{wang2018denseran} & $1.70\%$ & $8.44\%$  & $14.71\%$ & $19.51\%$ & $30.68\%$                       &                   DenseRAN\cite{wang2018denseran} & $0.20\%$ & $2.26\%$  & $7.89\%$ & $10.86\%$ & $24.80\%$  \\ 
                                      HDE\cite{cao2020zero} & $4.90\%$ & $12.77\%$  & $19.25\%$ & $25.13\%$ & $33.49\%$                       &                   HDE\cite{cao2020zero} & $7.48\%$ & $21.13\%$  & $31.75\%$ & $40.43\%$ & $51.41\%$  \\ 
                                      Stroke-based\cite{Chen2021ZeroShotCC} & $5.60\%$ & $13.85\%$  & $22.88\%$ & $25.73\%$ & $37.91\%$                      &                   Stroke-based\cite{Chen2021ZeroShotCC} & $7.03\%$ & $26.22\%$  & $48.42\%$ & $54.86\%$ & $65.44\%$ \\ 
                                      Ours & $\textbf{5.91\%}$ & $\textbf{14.35\%}$  & $\textbf{24.32\%}$ & $\textbf{30.17\%}$ & $\textbf{40.22\%}$                    &                  Ours & $\textbf{8.25\%}$ & $\textbf{32.24\%}$  & $\textbf{50.72\%}$ & $\textbf{57.13\%}$ & $\textbf{68.88\%}$  \\  \hline
\end{tabular}}
\end{center}
\end{table*}
\subsection{Experiments upon Pre-training}
In section 4, we propose two contrasting architectures in different forms of stroke sequence images for different tasks. To compare the performance of the two architectures on different forms of data sets, the first 500, 1000, 1500, 2000 and 2755 classes of 3755 printed Chinese characters are used as the training sets, and the last 1000 classes are used as the validation set. 
\begin{figure}[h]
\centering
	\subfloat[The first form of Chinese character stroke images.]{\label{fig:10a}\includegraphics[width = 8cm]{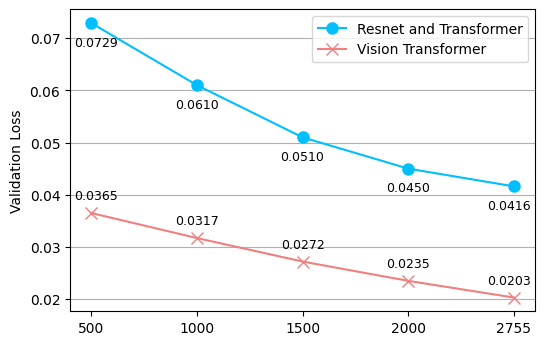}}
	\hfill
	\subfloat[The second form of Chinese character stroke images.]{\label{fig:10b}\includegraphics[width = 8cm]{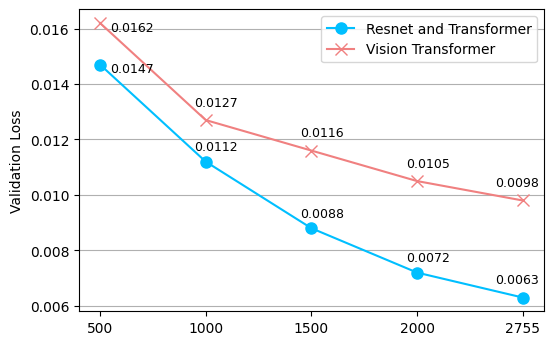}}
\caption{Validation loss on different forms of image sequence label and different architecture. Fig. \ref{fig:10a} is the validation loss on the first form, Fig. \ref{fig:10b} is the validation loss on the second form. }
\label{fig:10}
\end{figure}
The validation loss in different architectures and different forms of data sets is illustrated in Fig. \ref{fig:10}. Fig. \ref{fig:10a} shows the validation loss of two architectures on the first form of the datasets, and  Fig. \ref{fig:10b} shows the validation loss of two architectures on the second form of data sets.
First, we can see that the validation loss on the first form of data sets is significantly higher than on the second form. This is because each image in the second form only contains a single stroke and most of the pixels are zero. Information is much more sparse than in the first form. Therefore, it is easier to reconstruct.
Second, our SAE based on Resnet and Transformer predicts the next signal with teacher forcing strategy and, therefore, achieves smaller validation loss on second form of data sets.
But for the first form of data sets, the former performs worse than the latter. This is because we tackle the reconstruction problem as a seq2seq task. We take the stroke sequence of images as a sentence and each image as a word.
Since for the $t$-th image in the first form, it contains the preceding $t$-1 stroke information, which means that the previous images are redundant.
Thus, when predicting the next signal, the SAE architecture based on Resnet and Transformer will only focus on the last image and ignore other images, which is inconsistent with the idea of a seq2seq task. As a result, this architecture generalizes significantly worse than the SAE based on ViT.

Therefore, we conclude that these two architectures are suitable for different forms of datasets and different NLP tasks. For zero-shot recognition task, we adopt the SAE based on ResNet and Transformer pre-trained on the second form of data sets. While for the semantic enhancement task, it is more efficient for us to adopt the SAE based on ViT, since it can predict the image sequences dependent on only one input image without teacher forcing strategy.
Moreover, the first form of stroke image reflects the overall structural changes of Chinese character strokes,it has more global information than the second form and help our model utilize the global and local information of Chinese characters simultaneously.

\begin{figure*}[hb]
  \centering
  \center{\includegraphics[width=15cm]  {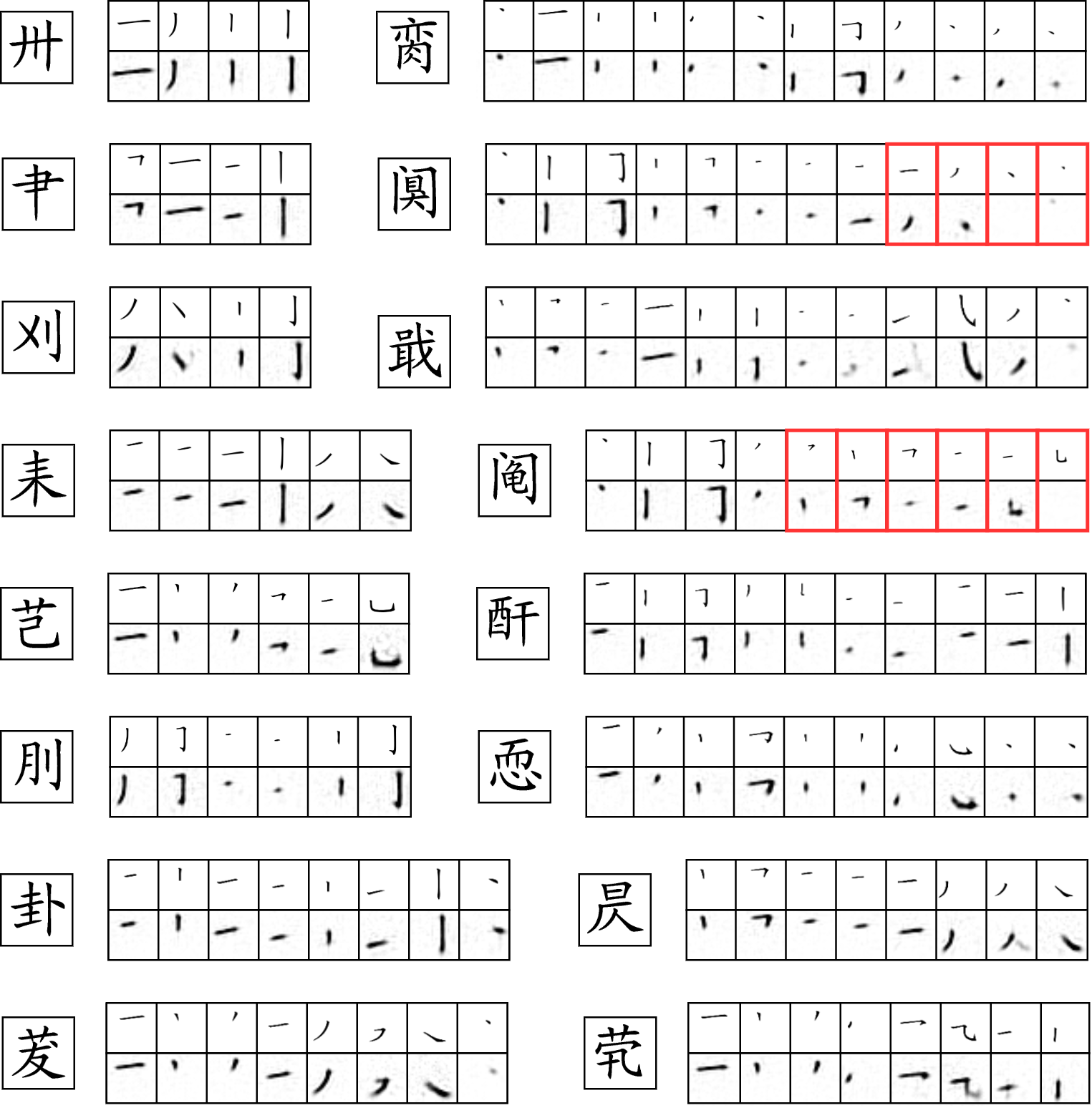}}
  \caption{Example results on validation datasets of second form pre-trained on SAE based on Resnet and Transformer. Each character has the original picture(above) and reconstructed picture(below). Those enclosed in red boxes represent some wrong reconstructions.}
  \label{fig:13}
\end{figure*}

We pre-train our SAE (Vision Transformer with the first form of datasets) on 3,755 Chinese Characters, and some example results has been illustrated in Fig. \ref{fig:2}. Fig. \ref{fig:13} gives more results pre-trained on the second form of datasets. The experiment demonstrates that after pre-training, our model can efficiently reconstruct the stroke sequence of unseen Chinese characters. This is because the encoder is capable of extracting semantic information from stroke images, and the decoder acquires the ability to predict the stroke sequences of Chinese characters.
Thus, our SAE can be applied not only to enrich word embeddings, but can also explore the structures of Chinese characters in zero-shot recognition.
We notice that for the characters '\begin{CJK*}{UTF8}{gbsn}阒\end{CJK*}' and '\begin{CJK*}{UTF8}{gbsn}阄\end{CJK*}', there are some wrong  reconstructions(highlighted in red boxes). This is because that both '\begin{CJK*}{UTF8}{gbsn}阒\end{CJK*}' and '\begin{CJK*}{UTF8}{gbsn}阄\end{CJK*}' are with semi-surrounded structures, and for these strokes other than the radical '\begin{CJK*}{UTF8}{gbsn}门\end{CJK*}', they are crowded together in tiny sizes and contain subtle differences,  which could make the reconstruction task even more challenging. 

\subsection{Experiments for Zero-Shot Chinese Character Recognition}
We employ the architecture based on ResNet and Transformer pre-trained on the second form of datasets for fine-tuning, as introduced in section 4.2. 
The pre-training model for different zero-shot tasks should be trained on the corresponding class of datasets, i.e., the first 500, 1000, 1500, 2000 and 2755 classes of 3,755 characters.

We compare our results with DenseRAN\cite{wang2018denseran}, HDE\cite{cao2020zero},  and stroke-based methods\cite{Chen2021ZeroShotCC}. The recognition accuracy of our method and other existing methods is shown in Table 1.
The results show that our method outperforms other methods on various kinds of data sets. Our pre-training method achieves greater improvements on larger datasets compared to stroke-based method. 
What is more encouraging is, our SAE is pre-trained on a small and concise dataset with printed Chinese character images, but boosts huge performance in zero-shot recognition on a large and complicated dataset consisting of handwritten Chinese character images.
To be specific, our SAE learns the stroke sequence of Chinese characters and spatial information of Chinese characters through masked image encoding, as a result, our SAE leverages the model to learn a better representation of handwritten Chinese characters at a higher semantic level. 

\section{Discussion: Intrinsic evaluation for word embeddings}
Our pre-trained SAE model is competent in word embeddings. We use the architecture based on Vision Transformer pre-trained on 3,755 printed Chinese character sequences to output word embeddings.
We notice that the output of the encoder is 25 vectors in 64 dimension, and each word vector corresponds to a spatial position of the image. 
A 1600($25\times64$) vector is too large for semantic embedding. For Chinese characters that have the same radical part at different locations, the embeddings of words cannot clearly show their relationship.
In addition, each output of the decoder corresponds to an image of Chinese character's stroke sequence, and therefore the middle layer of the decoder contains the stroke information of Chinese characters. 
For a Chinese character and its stroke number \textit{m}, the \textit{m}-th output corresponds exactly to its original image. 

\begin{figure}[h]
  \centering
  \center{\includegraphics[width=8cm]  {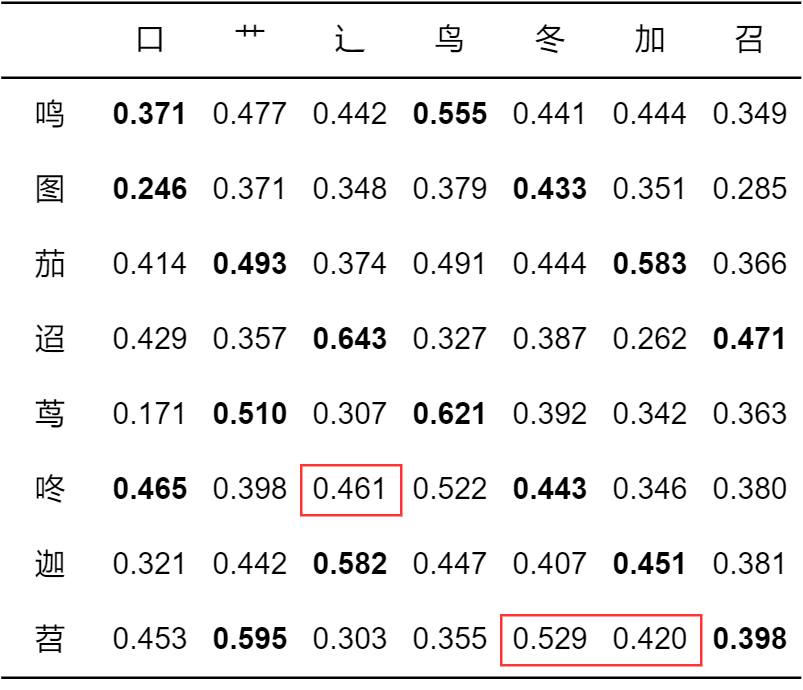}}
  \caption{The similarity between the Chinese characters and their radical parts. Bold numbers indicate the cosine similarity between Chinese characters and their corresponding radical components, and those enclosed in red boxes represent some outliers.}
  \label{fig:11}
\end{figure}

Based on the above analysis, we use the \textit{m}-th output patch of the third Transformer block of the decoder as its embedding vector, since it can reflect not only the stroke variation but also the overall structure of Chinese characters. To validate the representation ability of word embeddings of Chinese character structure, we calculate the word similarity of Chinese characters and their radical parts as an intrinsic evaluation. 
As shown in Figure \ref{fig:11}, the Chinese characters and their corresponding radical parts have a higher cosine similarity in that row and column. This illustrates that our encoder enhances the representation of Chinese characters with their abundant morphological information.

Moreover, we notice that the bold numbers in the first column do not reflect the strong relevance of Chinese characters and their radical parts. This is because all Chinese characters in the examples are composed of the same radical component '\begin{CJK*}{UTF8}{gbsn}口\end{CJK*}', except '\begin{CJK*}{UTF8}{gbsn}茑\end{CJK*}'.  
It can be seen that '\begin{CJK*}{UTF8}{gbsn}茑\end{CJK*}' has the lowest cosine similarity in the first column, which is consistent with the above observation.

  \begin{figure}[htbp]
  \centering
  \center{\includegraphics[width=8cm]  {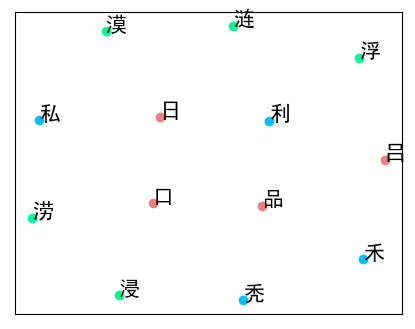}}
  \caption{T-SNE visualization of embeddings learned from SAE architecture(different colors represent different classes).}
  \label{fig:12}
\end{figure}
We use the t-SNE visualization\cite{tsne}  for the 256-dim embeddings of some Chinese characters. The K-means clustering algorithm\cite{kmeans} with cosine distance is applied to divide these characters into three categories, and each category is marked with the same color. As shown in Fig. \ref{fig:12},  we can see that these Chinese characters in the same category really have some common radical and stroke parts. Thus, the clustering results show great potential for our word embedding ability in other Chinese NLP-related applications, such as semantic enhancement and style transfer (e.g., from handwritten to printed and vice versa).
Our discussion of word embeddings from SAE architecture only scratches the surface of the Chinese character structure in semantic enhancement. How to apply SAE to other Chinese NLP tasks is beyond the scope of this work, but worthy of further exploration.

\section{Conclusion}
In this paper, we propose stroke-based autoencoders to learn the structural information of Chinese characters.
we propose two contrasting SAE architectures on different forms of data sets, which are employed for the zero-shot handwritten Chinese character recognition task and semantic enhancement task.
Through combining our pre-training method and existing method, the proposed approach manages to achieve improved recognition accuracy results over existing methods and boost significant performance on a large and complicated dataset by pre-training on a small and concise dataset.
The experimental results on the benchmark datasets compared with existing methods validate that our SAE can effectively deal with the zero-shot Chinese character recognition problem. 
Among future directions, one particularly interesting extension is to enrich the Chinese word embeddings and to facilitate other related Chinese NLP tasks. . 
\section*{Acknowledgement}
This work is partially supported by the National Natural Science Foundation of China (61573012), China Scholarship Council (201906955038), and National innovation and entrepreneurship training program for college students (3120400002300).




\balance

\end{document}